\documentclass{article}

\usepackage{iclr2026_conference,times}
\iclrfinalcopy
\usepackage[T1]{fontenc}
\usepackage{microtype}
\usepackage{enumitem}
\setlist{nosep,leftmargin=*}

\usepackage{amsmath,amssymb,amsfonts}
\usepackage{mathtools}
\usepackage{bm}

\usepackage{graphicx}
\usepackage{subcaption}
\usepackage{xcolor}

\usepackage{booktabs}
\usepackage{multirow}
\usepackage{array}
\usepackage{makecell}

\usepackage{algorithm}
\usepackage{algorithmic}

\usepackage[breaklinks=true]{hyperref}
\hypersetup{
    colorlinks=true,
    linkcolor=blue!70!black,
    citecolor=blue!50!black,
    urlcolor=blue!70!black,
    pdftitle={Contextual Evaluation for Materials Benchmarking},
    pdfauthor={Lu, Tripathi, Shteyn, Ganguly, French, Chaudhary, Wu},
}

\usepackage{xcolor}
\usepackage[normalem]{ulem} 
\usepackage[capitalize,noabbrev]{cleveref}

\usepackage{tikz}
\usetikzlibrary{arrows.meta,positioning,calc,shapes.symbols}

    \newcommand{\yw}[1]{
    \noindent\textbf{[YW:}- \textcolor{red}{#1}]}

\newcommand{\bi}{\begin{itemize}}
\newcommand{\ei}{\end{itemize}}

\newcommand{\eat}[1]{}

\newcounter{example}
\renewcommand{\theexample}{\arabic{example}}

\newcommand{\stitle}[1]{\vspace{.5ex}\noindent{\bf #1}}
\newcommand{\eetitle}[1]{\vspace{0.8ex}\noindent{\underline{\em #1}}}

\newcommand{\sstab}{\rule{0pt}{8pt}\\[-2.4ex]}

\title{Context Determines Optimal Architecture \\
in Materials Segmentation}

\author{Mingjian Lu\textsuperscript{1} \quad
Pawan K. Tripathi\textsuperscript{1} \quad
Mark Shteyn\textsuperscript{1} \quad
Debargha Ganguly\textsuperscript{1} \quad \\
\textbf{Roger H. French}\textsuperscript{1} \quad
\textbf{Vipin Chaudhary}\textsuperscript{1} \quad
\textbf{Yinghui Wu}\textsuperscript{1}\\[0.5ex]
\textsuperscript{1}Case Western Reserve University, Cleveland, Ohio, USA\\
\texttt{\{mxl1171, pawan.tripathi, mis60, debargha,\}@case.edu}\\
\texttt{\{roger.french, vipin, yxw1650\}@case.edu}\\
\url{https://github.com/cwru-sdle/materials-data-segmentation-benchmark}
}

\begin{document}

\maketitle

\begin{abstract}
Segmentation architectures are typically benchmarked on single imaging modalities, obscuring deployment-relevant performance variations: an architecture optimal for one modality may underperform on another. We present a cross-modal evaluation framework for materials image segmentation spanning SEM, AFM, XCT, and optical microscopy. Our evaluation of six encoder-decoder combinations across seven datasets reveals that optimal architectures vary systematically by context: UNet excels for high-contrast 2D imaging while DeepLabv3+ is preferred for the hardest cases. The framework also provides deployment feedback via out-of-distribution detection and counterfactual explanations that reveal which microstructural features drive predictions. Together, the architecture guidance, reliability signals, and interpretability tools address a practical gap in materials characterization, where researchers lack tools to select architectures for their specific imaging setup or assess when models can be trusted on new samples.
\end{abstract}

\vspace{-1em}
\section{Introduction}
\label{sec:introduction}

Quantitative microstructural analysis underpins materials discovery across manufacturing, energy, and aerospace sectors~\citep{holm2020overview,bostanabad2018computational,choudhary2022recent}. Deep learning has enabled automated segmentation for tasks ranging from pore detection in additive manufacturing~\citep{stan2020optimizing} to defect identification in electron microscopy~\citep{roberts2019deep,azimi2018advanced}. However, architectures that excel on one imaging modality often underperform on another: in our benchmarks, best-performing models exhibit IoU $\in [0.47, 0.95]$, a 48 percentage-point spread that persists despite identical training protocols. This variation indicates that inductive biases interact non-trivially with modality-specific statistics. Relative model rankings that depend on evaluation domain violate the implicit independent and identically distributed (i.i.d.) assumption underlying standard benchmarks, which report aggregate performance rather than deployment guidance for specific modalities and material systems~\citep{ma2020image,pokharel2024comparative}.

As characterization pipelines become automated~\citep{hernandezLPBFHighThroughputData2024,ciardiMaterialsDataScience2024}, this gap creates deployment risks. Segmentation models exhibit high-confidence errors under covariate shift, producing semantically plausible but incorrect masks on out-of-distribution samples (due to batch variation, preparation differences, or imaging drift), and these errors propagate undetected into downstream analyses: flawed porosity fractions, grain size distributions, or corrosion rates~\citep{kalutotageAutomatedImageSegmentationc}. Addressing this requires context-conditional evaluation that provides both architecture guidance and reliability signals for deployment.


\begin{figure}[tb!]
\centering
\begin{subfigure}[t]{0.48\columnwidth}
\centering
\includegraphics[width=\linewidth]{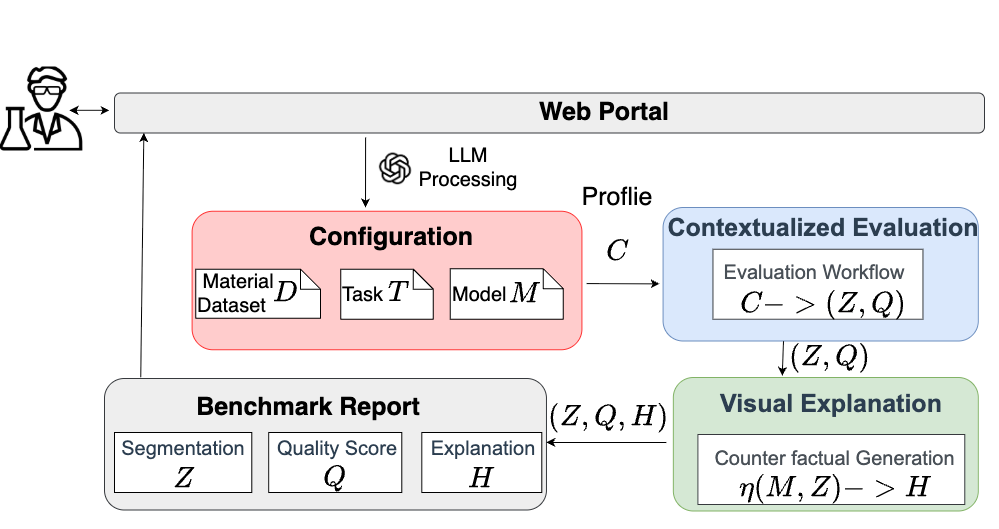}
\caption{Framework overview: three modules for cross-modal configuration, quality feedback, and expert knowledge integration.}
\label{fig:architecture}
\end{subfigure}
\hfill
\begin{subfigure}[t]{0.48\columnwidth}
\centering
\includegraphics[width=\linewidth]{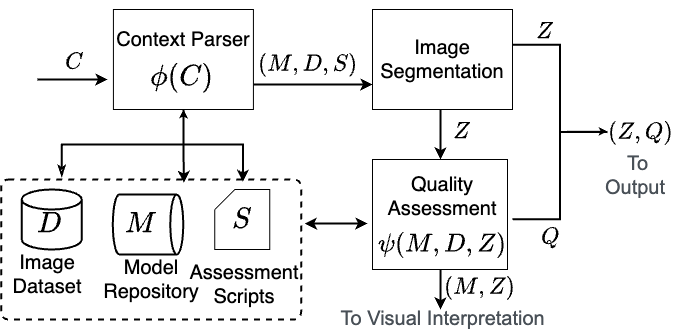}
\caption{Evaluation workflow: configuration-driven data/model selection, segmentation, quality assessment, and report generation.}
\label{fig:eval_workflow}
\end{subfigure}
\caption{(a) The three-component evaluation framework maps heterogeneous inputs to standardized profiles, produces segmentation masks with reliability signals, and generates interpretability heatmaps. (b) The quality-controlled evaluation workflow executes this pipeline.}
\label{fig:framework_overview}
\vspace{-2ex}
\end{figure}

We present a configurable evaluation framework addressing this gap through three capabilities:
\begin{enumerate}
    \item \textbf{Cross-modal architecture comparison.} Systematic evaluation testing whether architecture rankings are stable across imaging modalities, or whether optimal inductive biases are context-dependent, using six encoder-decoder combinations: two encoders (ResNet50~\citep{he2016deep}, SE-ResNeXt101~\citep{hu2018squeeze}) and three decoders (UNet~\citep{ronneberger2015u}, DeepLabv3~\citep{chen2017rethinking}, DeepLabv3+~\citep{chen2018encoder}).

    \item \textbf{Cross-modal architecture comparison.} Systematic evaluation validating how architectural inductive biases map to specific materials imaging constraints (e.g., surface vs.\ volumetric), using six encoder-decoder combinations: two encoders (ResNet50~\citep{he2016deep}, SE-ResNeXt101~\citep{hu2018squeeze}) and three decoders (UNet~\citep{ronneberger2015u}, DeepLabv3~\citep{chen2017rethinking}, DeepLabv3+~\citep{chen2018encoder}).

    \item \textbf{Deployment quality control.} Out-of-distribution detection~\citep{gangulyForteFindingOutliers2024a} that identifies distribution shift before it manifests as segmentation error, converting silent failures into actionable warnings for scientific workflows~\citep{lu2023image,10386478}.

    \item \textbf{Expert-interpretable explanations.} Counterfactual analysis~\citep{vermeire2022explainable} that identifies the minimal image modifications that change predictions, revealing which microstructural features are decision-critical for validation against domain knowledge~\citep{modarres2017neural}.
\end{enumerate}

Evaluating six encoder-decoder combinations across seven datasets, we find systematic context-dependence: decoder architecture interacts with task difficulty, suggesting that architectural inductive biases must be matched to image characteristics. UNet dominates high-contrast 2D tasks (SEM, AFM, Optical), while DeepLabv3+ is preferred for the hardest cases (fractography SEM and XCT stress-corrosion) where multi-scale context is critical.

\vspace{-1em}
\section{Methodology}
\label{sec:methodology}

To address context-dependent architecture selection, we introduce a three-component evaluation framework for materials image segmentation across heterogeneous modalities (Figure~\ref{fig:architecture}). At a high level, the framework takes dataset specifications, task descriptions, and model choices as input, and produces segmentation masks, reliability signals indicating potential distribution shift, and interpretability heatmaps for expert validation.


More formally, the \emph{Cross-Modal Configuration Module} standardizes heterogeneous input specifications (dataset $\mathcal{D}$, task $T$, model $M$) into a configuration $C$ within configuration space $\mathcal{C}$, ensuring consistent evaluation semantics across modalities. The \emph{Quality Feedback Module} takes $C$ to produce segmentation masks $Z$ and reliability signals $Q$, while the \emph{Expert Knowledge Integration Module} operates on $(M, Z)$ to generate interpretability heatmaps $H$. These outputs $(Z, Q, H)$ are summarized in a structured feedback report.

These components induce a feedback process wherein reliability signals $Q$ expose distribution shift, prompting configuration revision, while explanations $H$ enable expert validation of learned decision boundaries against domain priors. The loop terminates when $Q$ indicates acceptable calibration and $H$ aligns with domain expectations.

The evaluation pipeline thus factors into three stages: configuration $C$ determines $(D, M)$; inference and quality assessment yield $(Z, Q)$; and the interpreter produces $H$ conditioned on expert-specified regions. The resulting tuple $(Z, Q, H)$ forms a complete evaluation record enabling reproducible, context-specific analysis.

\stitle{Cross-Modal Configuration Module}.
\label{sec:configurable}
The configuration module implements $\pi$ by aggregating dataset specifications $\mathcal{D}$ with feature representations $D_F$ and task descriptors $T$, producing a configuration profile $C \in \mathcal{C}$ that serves as the canonical input to downstream evaluation. Profiles are standardized JSON documents (Table~\ref{tab:config_profile}) specifying three categories: 

\textbf{Dataset Configuration.} Specifies properties of datasets $D$, including (1) imaging modalities (e.g., SEM, AFM, XCT, Optical), (2) material systems (e.g., metals, polymers, semiconductors), and (3) auxiliary metadata and provenance such as materials ontologies and experiment descriptors \citep{rajamohan2025materials}. 

\textbf{Task Configuration.} Contains (1) natural language statements of a task description $T$ that describes segmentation goals (e.g., ``Identify crystal impingement sites''), and (2) 
performance measures \citep{cho2024weighted} and scripts for assessing segmentation quality and model performance.  

\textbf{Model Specification.} Declares the segmentation model $M = F_2 \circ F_1$ as the composition of an encoder $F_1$ and decoder $F_2$. In this study, encoders include ResNet50 \citep{he2016deep} and SE-ResNeXt101 \citep{hu2018squeeze,xie2017aggregated}, while decoders include UNet \citep{ronneberger2015u}, DeepLabv3 \citep{chen2017rethinking}, and DeepLabv3+ \citep{chen2018encoder}.

\begin{table}[tb!]
\centering
\caption{Configuration profile schema defining the evaluation space $\mathcal{C}$ for cross-modal benchmarking.}
\label{tab:config_profile}
\footnotesize
\setlength{\tabcolsep}{4pt}
\renewcommand{\arraystretch}{1.05}
\begin{tabular}{@{}llll@{}}
\toprule
\textbf{Category} & \textbf{Property} & \textbf{Values} & \textbf{Description} \\
\midrule
\multirow{3}{*}{Dataset}
  & Modality & SEM, AFM, XCT, Optical & Resolution: sub-nm to $\mu$m~\citep{holm2020overview} \\
  & Material & Metals, Polymers, Semicond. & Microstructural features~\citep{choudhary2022recent} \\
  & Metadata & Ontological descriptors & Provenance tracking~\citep{rajamohan2025materials} \\
\midrule
\multirow{3}{*}{Task}
  & Objective & ``Identify grain boundaries'' & Natural language specification \\
  & Metrics & IoU, Dice, pixel accuracy & Quality measures~\citep{cho2024weighted} \\
  & QC signals & AUROC, $F_1$, precision & OOD indicators~\citep{gangulyForteFindingOutliers2024a} \\
\midrule
\multirow{2}{*}{Model}
  & Encoders & ResNet50, SE-ResNeXt101 & Backbone~\citep{he2016deep,hu2018squeeze} \\
  & Decoders & UNet, DeepLabv3(+) & Seg.\ head~\citep{ronneberger2015u,chen2018encoder} \\
\bottomrule
\end{tabular}

\end{table}

The configuration profile $C$ specifies the Cartesian product of candidate encoders and decoders for systematic evaluation. While $C$ can be authored directly as structured input (JSON), future extensions could compile natural language specifications into $C$ via NLP pipelines, further supporting reproducibility across material-imaging-task contexts. 

\stitle{Quality-Controlled Evaluation with Feedback Signals}.
\label{sec:evaluation}
Given configuration $C$, this module executes the evaluation workflow (Figure~\ref{fig:eval_workflow}) via two components:
\textbf{(1) Query parser} $\phi: \mathcal{C} \to \mathcal{Q} \times \mathcal{J}$: A deterministic compilation function that maps configuration $C$ to an executable query set $\mathcal{Q}$ and job specification $\mathcal{J}$, which (i) loads the dataset $D$ and its representation $D_F$, (ii) retrieves encoder $F_1$ and decoder $F_2$ specifications and assembles the segmentation model $M = F_2 \circ F_1$, and (iii) generates training and evaluation scripts.
\textbf{(2) Performance evaluator} $\psi: M \times \mathcal{D} \times \mathcal{C} \to (Z, Q)$: Computes the segmentation map $Z = M(D)$ and derives quality feedback $Q = (q_\text{AUROC}, q_{F_1}, q_\text{prec})$ via the OOD detection procedure specified in $C$~\citep{gangulyForteFindingOutliers2024a}, which utilizes self-supervised embeddings and typicality estimation to flag outliers, where $Q$ quantifies model calibration under potential covariate shift and indicates when retraining or domain adaptation may be warranted.

\stitle{Expert Knowledge Integration through Counterfactual Feedback}.
\label{sec:interpretability}
Deployment in scientific workflows requires attribution of predictions to input features, enabling validation that learned representations encode domain-relevant structure. Our framework generates counterfactual explanations focused on designated areas of interest via the interpreter function $\eta: M \times Z \to H$. Given an input image $D$, its prediction $Z = M(D)$, and a user annotation mask $A$ marking the region of interest, we construct the perturbed input $D^{(i,j)} = D + \delta_{ij}$ for each spatial location $(i,j)$, where $\delta_{ij}$ is a localized perturbation kernel, and compute the counterfactual prediction $Z^{(i,j)} = M(D^{(i,j)})$. The influence heatmap $H$ is computed as
\[
H_{i,j} = \left\lVert (Z \odot A) - (Z^{(i,j)} \odot A) \right\rVert_{1},
\]
where $H_{i,j}$ measures the $\ell_1$ sensitivity of predictions within $A$ to perturbations at location $(i,j)$. This formulation restricts attribution to expert-specified structures, yielding a spatially-localized sensitivity map that identifies which input regions exert maximal influence on predictions within the annotated area.

\section{Experiments and Case Analysis}
\label{sec:experiments}
\vspace{-0.5em}

\stitle{Experimental Setup}.
We conduct a systematic evaluation across encoder--decoder combinations to assess performance across materials research scenarios and tasks. All segmentation models $M$ use ImageNet-pretrained encoders and identical training settings: batch size 16, learning rate $10^{-4}$ with cosine annealing, and early stopping with patience 20. Code is available at \citep{lu2025materials_benchmark} for reproducibility.

\eetitle{Configurations}. We adopt the configuration schema and example values in Table~\ref{tab:config_profile}. Dataset details and additional settings to aid reproducibility are reported in \citep{lu2025materials_benchmark}.

\stitle{Experimental Results}.
We report our findings below.
\vspace{-1em}
\begin{table}[!htbp]
\centering
\caption{Cross-modal segmentation performance. SE-ResNeXt101 is the optimal encoder across all configurations; we report only the best-performing decoder. Difficulty reflects intrinsic task complexity rather than achieved IoU. (Full results are in Table \ref{tab:full_architecture_results}.)}
\label{tab:comprehensive_results}
\resizebox{\columnwidth}{!}{%
\begin{tabular}{@{}lllllcc@{}}
\toprule
\textbf{Modality} & \textbf{Dataset} & \textbf{Material} & \textbf{Task} & \textbf{Difficulty} & \textbf{IoU} & \textbf{Dice} \\
\midrule
AFM & Crystallites~\citep{lu2023image} & Fluoroelastomer & Spherulite/Lamellae & Moderate & \textbf{0.901} & 0.961 \\
\midrule
\multirow{3}{*}{SEM}
  & Contact Corr. & Si Solar Cell & Corroded/Intact & Easy & \textbf{0.951} & 0.962 \\
  & Fractography$^\dagger$ & Ti-6Al-4V & Defects & Hard & \textbf{0.636} & 0.490 \\
  & Carbon SEM~\citep{pokharel2024comparative} & Carbon & Carbon regions & Moderate & \textbf{0.762} & 0.621 \\
\midrule
\multirow{2}{*}{XCT}
  & Al Pitting~\citep{kalutotageAutomatedImageSegmentationc} & Al Wire & Pitting corrosion & V.~Hard & \textbf{0.752} & 0.570 \\
  & Al-Mg-Si SC$^\dagger$~\citep{ciardiMaterialsDataScience2024} & Al-Mg-Si & Stress corrosion & V.~Hard & \textbf{0.472} & 0.550 \\
\midrule
Optical & L-PBF~\citep{hernandezLPBFHighThroughputData2024} & 316L Steel & Melt pool/Spatter & Hard & \textbf{0.652} & 0.628 \\
\bottomrule
\multicolumn{7}{@{}l@{}}{\scriptsize $^\dagger$DeepLabv3+ decoder; all others use UNet.}
\end{tabular}%
}
\end{table}
\vspace{-1em}

\eetitle{Exp-1: Cross-Modal Architecture Recommendations.}
Table~\ref{tab:comprehensive_results} reports IoU and Dice scores across seven datasets spanning four imaging modalities. Our evaluation reveals three natural use-case categories, each sharing common data characteristics (high-contrast boundaries, temporal dynamics, or multi-scale volumetric structure) that benefit from consistent architecture choices. We summarize the findings below.

\sstab
\textit{Use Case 1: High-Contrast Surface Characterization.}
For datasets with sharp boundaries and high contrast (AFM, surface SEM), SE-ResNeXt101 paired with UNet achieves IoU above 0.90, including 0.951 for semiconductor contact corrosion and 0.901 for polymer crystallite analysis. UNet's skip connections preserve fine spatial gradients critical for accurate boundary delineation, while SE-ResNeXt101's channel attention amplifies contrast at material interfaces.

\sstab
\textit{Use Case 2: Dynamic Process Monitoring.}
For real-time characterization tasks (L-PBF additive manufacturing, Carbon SEM), the same SE-ResNeXt101 + UNet combination yields IoU between 0.65--0.76. The lower performance vs.\ Use Case 1 reflects challenges like temporal variation in melt-pool morphology rather than architectural mismatch. The attention mechanism in SE-ResNeXt101 helps emphasize transient, high-contrast regions, while UNet's lower latency better suits high-throughput in-situ monitoring.

\sstab
\textit{Use Case 3: Multi-Scale Volumetric Analysis.}
For 3D volumetric tasks requiring multi-scale reasoning (XCT corrosion, fractography), switching to DeepLabv3+ as the decoder improves performance, achieving IoU of 0.47--0.75. This includes challenging cases like stress corrosion cracking (IoU 0.472), pitting corrosion (IoU 0.752), and fractography defects (IoU 0.636). DeepLabv3+'s atrous spatial pyramid pooling captures defect patterns across multiple scales, and its boundary refinement module better handles the irregular morphologies of sub-surface crack networks.

\eetitle{Summary.} Our contextualized evaluation reveals three key findings that static, aggregate benchmarks would obscure:
\begin{enumerate}[leftmargin=*,itemsep=2pt,topsep=2pt]
\item \textbf{No single architecture dominates.} While SE-ResNeXt101 consistently outperforms ResNet50 as an encoder, optimal decoder choice depends on task characteristics: UNet excels for high-contrast surface tasks, while DeepLabv3+ is preferred for volumetric analysis.
\item \textbf{Performance is context-sensitive.} The same architecture (SE-ResNeXt101 + UNet) achieves IoU of 0.901 on crystallite analysis but only 0.652 for manufacturing monitoring, a 25 percentage-point spread that would be masked by aggregate reporting.
\item \textbf{Task difficulty guides decoder selection.} Easy and moderate tasks favor UNet; hard and very hard tasks, especially those involving complex morphologies or volumetric defects, benefit from DeepLabv3+'s multi-scale context.
\end{enumerate}

\begin{figure}[tb!]
\centering
\includegraphics[width=0.9\columnwidth]{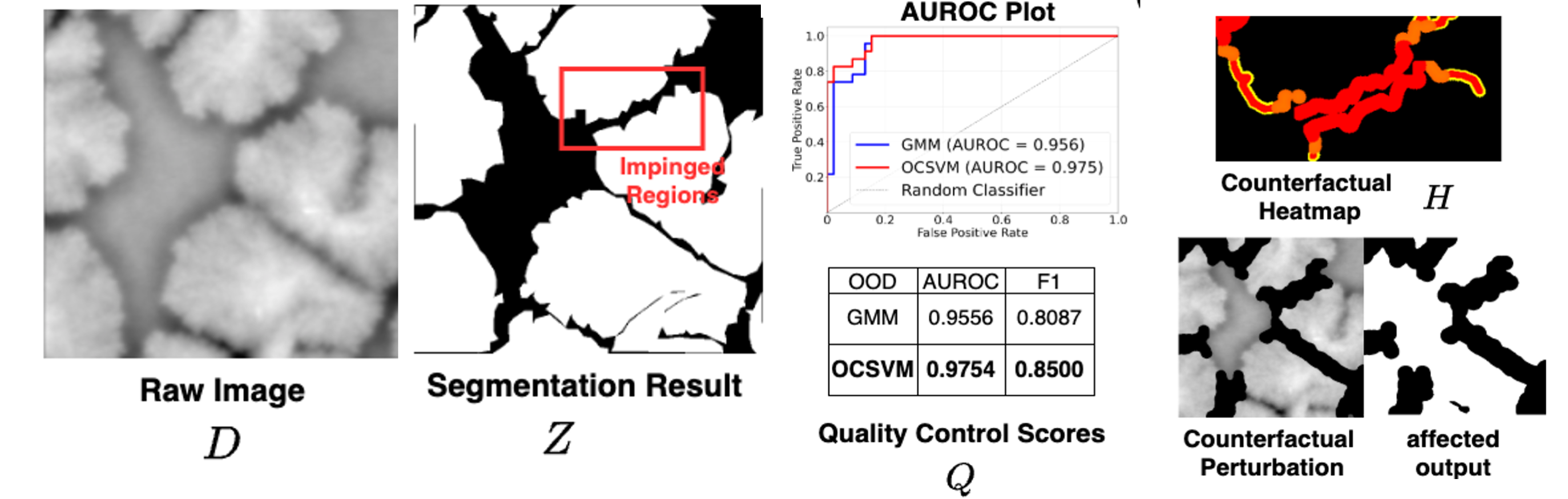}
\vspace{-1ex}
\caption{{\small Complete feedback loop demonstration for AFM crystallite analysis: (a) Multi-channel input $D$, (b) Segmentation output $Z$, (c) Quality feedback signals $Q$ (AUROC, F1) indicating deployment readiness, (d) Interpretable feedback $H$ enabling expert validation of decision-relevant regions. }}
\label{fig:afm_case}
\vspace{-3ex}
\end{figure}

\eetitle{Exp-2: End-to-End Feedback Loop Demonstration.}
We illustrate a complete feedback cycle for AFM crystallite analysis, from task specification to expert validation:

\textbf{Step 1 (Configuration):} The user specifies the task (``Identify crystal impingement sites in polycrystalline fluoroelastomer''), yielding a configuration for AFM multi-channel data (Figure~\ref{fig:afm_case}a).

\textbf{Step 2 (Model Evaluation):} The framework evaluates all encoder-decoder combinations. SE-ResNeXt101 + UNet achieves the best performance with IoU = 0.901 and Dice = 0.961 (Figure~\ref{fig:afm_case}b).

\textbf{Step 3 (Quality Assessment):} The OOD detector produces deployment-readiness scores: AUROC $>$ 0.95 and F1 $>$ 0.80 under both GMM and OCSVM estimators (Figure~\ref{fig:afm_case}c). Since AUROC exceeds the 0.95 threshold, the model is flagged as ready for deployment.

\textbf{Step 4 (Report Generation):} The framework consolidates segmentation masks, quality scores, and interpretability heatmaps into a structured evaluation report (Figure~\ref{fig:afm_case}a--d).

This walkthrough demonstrates how the framework enables an iterative workflow: from task specification to architecture selection to quality assessment to expert validation, supporting rapid iteration for accelerated materials characterization.

\eat{
\eetitle{Exp-3: Visual Counterfactual Explanation}. The interpretability function $\eta(M,Z) \rightarrow H$ generates heatmaps $H$ (Figure~\ref{fig:afm_case}d) revealing that the optimal model focuses on crystal impingement sites as specified in task $T$. The heatmaps highlight the red-circled impingement regions from the original AFM image (Figure~\ref{fig:afm_case}a), where crystallites meet and form boundaries in the segmentation mask (Figure~\ref{fig:afm_case}b). Through systematic modification of input features, we identify that the model prioritizes topographic variations and phase discontinuities at these impingement sites, demonstrating scientifically relevant decision-making aligned with the crystallographic analysis objectives.
\yw{Make this more specific as discussed. }
}

\eetitle{Exp-3: Expert Knowledge Integration.}
The interpretability heatmaps (Figure~\ref{fig:afm_case}d) confirm that the model focuses on scientifically meaningful features rather than spurious correlations. High-influence regions (red/orange) correspond precisely to crystal impingement sites, where crystallites meet and form boundaries, matching expert annotations in the original image (Figure~\ref{fig:afm_case}a) and boundary regions in the segmentation output (Figure~\ref{fig:afm_case}b).

Perturbation analysis validates this interpretation: blurring the high-influence regions significantly degrades predictions, while perturbing background regions has minimal effect. This confirms that the model encodes sensitivity to topographic gradients and phase discontinuities at grain boundaries, the physically meaningful features for crystallographic analysis, rather than dataset-specific artifacts. Such validation provides confidence that the chosen architecture generalizes appropriately for boundary characterization tasks.
\section{Conclusions}
\label{sections/conclusion}
We present a cross-modal evaluation framework for materials image segmentation that provides architecture guidance, deployment-readiness signals, and interpretable explanations. Our evaluation across seven datasets reveals that optimal architectures vary systematically by context: UNet excels for high-contrast surface imaging while DeepLabv3+ is preferred for volumetric analysis. By combining quality control with expert-interpretable feedback, the framework supports reliable deployment in scientific workflows where silent failures carry real cost. %

\section*{Acknowledgments}
This work is supported by the Department of Energy's National Nuclear Security Administration under Award Number DE-NA0004104 (MDS\textsuperscript{3}-COE), and NSF Grant 2117439.We extend our sincere thanks to Christine Orme (Lawrence Livermore National Laboratory, LLNL) for providing the AFM dataset; Jean-Baptiste Forien, Brian Giera, and Ibo Matthews (LLNL) for the L-PBF HSC dataset; John Lewandowski (CWRU) for the Fractography and Al-Mg-Si SC XCT datasets; Max Liggett and Kris Davis (University of Central Florida); and Philip Noell (Sandia National Laboratories) for the Al Pitting XCT dataset.
\clearpage
\bibliography{references}
\bibliographystyle{iclr2026_conference}
\clearpage
\appendix
\section{Appendix}

\subsection{Additional Experiment Details to aid reproducibility}
\stitle{Details of Datasets}. 
Our framework is evaluated on seven diverse materials datasets spanning multiple imaging modalities, material systems, and segmentation challenges. Each dataset represents real-world materials characterization scenarios with expert-validated ground truth annotations.

\textbf{Imaging Modalities:} We cover four primary characterization techniques: (1) \textit{Scanning Electron Microscopy (SEM)} providing high-resolution surface and cross-sectional analysis with submicron scale features; (2) \textit{Atomic Force Microscopy (AFM)} offering multi-channel topographical, phase, and amplitude information at 512$\times$512 resolution; (3) \textit{X-ray Computed Tomography (XCT)} enabling 3D volumetric analysis of internal structures at 1--2$\mu$m/voxel resolution; and (4) \textit{Optical microscopy} for dynamic process monitoring at high temporal resolution ($\sim$1kHz).

\textbf{Material Systems:} The datasets encompass diverse material classes critical to various applications: metals including 316L stainless steel for additive manufacturing, Ti-6Al-4V for aerospace applications, and aluminum alloys for structural components; polymers such as fluoroelastomers (Kel-F, FK800) for chemical resistance applications; semiconductors including silicon solar cells for photovoltaic systems; and carbon-based materials for advanced applications.

\textbf{Segmentation Challenges:} Tasks span the full spectrum of materials characterization needs: (1) \textit{Microstructure analysis} including crystalline spherulites and lamellar structures in polymers; (2) \textit{Defect detection} encompassing manufacturing defects in fractography, keyhole porosity, and lack-of-fusion defects; (3) \textit{Corrosion assessment} covering contact corrosion in solar cells, pitting corrosion in aluminum, and stress corrosion cracking; and (4) \textit{Dynamic phenomena} such as melt pool and spatter detection in laser powder bed fusion processes.

\textbf{Difficulty Classification:} Each dataset is classified into five difficulty levels based on segmentation complexity: \textit{Easy} cases feature high contrast and clear boundaries (e.g., contact corrosion in SEM); \textit{Moderate} cases involve well-defined structures with some texture complexity (e.g., AFM crystallites); \textit{Hard} cases require distinguishing subtle features or handling class imbalance (e.g., fractography defects); and \textit{Very Hard} cases involve complex 3D structures with low contrast and significant noise (e.g., stress corrosion in XCT). This classification enables systematic evaluation of architecture performance across varying complexity levels.

\textbf{Evaluation metrics.}
Segmentation performance is evaluated using Intersection over Union (IoU) \citep{everingham2010pascal}, Dice coefficient \citep{milletari2016v}, and a weighted variant of IoU to account for class imbalance and structural heterogeneity across datasets.

Let $D$ denote an input image, $M$ a segmentation model, and $Z = M(D)$ the predicted segmentation map.
Let $Z^{\ast}$ denote the corresponding ground-truth segmentation.

The standard Intersection over Union (IoU) is defined as
\begin{equation}
\mathrm{IoU}
=
\frac{| Z \cap Z^{\ast} |}{| Z \cup Z^{\ast} |}.
\end{equation}

The Dice coefficient is computed as
\begin{equation}
\mathrm{Dice}
=
\frac{2 | Z \cap Z^{\ast} |}{| Z | + | Z^{\ast} |},
\end{equation}
which emphasizes overlap consistency and is commonly used alongside IoU in materials image segmentation.

To mitigate the dominance of large homogeneous regions and better reflect segmentation quality on small or critical structures, we additionally consider the weighted Intersection over Union (wIoU)~\citep{cho2024weighted}.
Let $Z_i$ and $Z^{\ast}_i$ denote the predicted and ground-truth labels at spatial index $i$, and let $w_i$ be a non-negative weight.
The weighted IoU is defined as
\begin{equation}
\mathrm{wIoU}
=
\frac{\sum_i w_i \, | Z_i \cap Z^{\ast}_i |}
{\sum_i w_i \, | Z_i \cup Z^{\ast}_i |}.
\end{equation}

In our experiments, weights $w_i$ are derived from class-frequency statistics to emphasize underrepresented structures, ensuring that performance comparisons remain robust across datasets with varying class imbalance and segmentation difficulty.

\textbf{Full encoder--decoder evaluation results.}
To ensure transparency in architecture selection and to support the context-dependent conclusions reported in the main text, Table~\ref{tab:full_architecture_results} reports the complete segmentation performance for all evaluated encoder--decoder combinations across datasets.
These results constitute the full empirical basis from which the best-performing configurations summarized in Table~2 are selected.

\begin{table*}[t]
\centering
\caption{Complete encoder--decoder segmentation results across all datasets.
All evaluated configurations are reported.
For each dataset, the best-performing configuration (based on IoU) is highlighted in bold.
}
\label{tab:full_architecture_results}
\setlength{\tabcolsep}{4.5pt}
\renewcommand{\arraystretch}{1.08}
\footnotesize
\begin{tabular}{l l l c c}
\toprule
Dataset & Encoder & Decoder & IoU & Dice \\
\midrule
\multirow{6}{*}{L-PBF}
 & ResNet50 & UNet & 0.578 & 0.551 \\
 & ResNet50 & DeepLabv3 & 0.342 & 0.387 \\
 & ResNet50 & DeepLabv3+ & 0.471 & 0.597 \\
 & SE-ResNeXt101 & UNet & \textbf{0.652} & \textbf{0.628} \\
 & SE-ResNeXt101 & DeepLabv3 & 0.353 & 0.395 \\
 & SE-ResNeXt101 & DeepLabv3+ & 0.494 & 0.612 \\
\midrule
\multirow{6}{*}{Fractography}
 & ResNet50 & UNet & 0.593 & 0.516 \\
 & ResNet50 & DeepLabv3 & 0.568 & 0.487 \\
 & ResNet50 & DeepLabv3+ & 0.576 & 0.501 \\
 & SE-ResNeXt101 & UNet & 0.627 & 0.548 \\
 & SE-ResNeXt101 & DeepLabv3 & 0.636 & 0.450 \\
 & SE-ResNeXt101 & DeepLabv3+ & \textbf{0.636} & \textbf{0.490} \\
\midrule
\multirow{6}{*}{AFM Crystallites}
 & ResNet50 & UNet & 0.897 & 0.902 \\
 & ResNet50 & DeepLabv3 & 0.808 & 0.843 \\
 & ResNet50 & DeepLabv3+ & 0.894 & 0.937 \\
 & SE-ResNeXt101 & UNet & \textbf{0.901} & \textbf{0.961} \\
 & SE-ResNeXt101 & DeepLabv3 & 0.892 & 0.959 \\
 & SE-ResNeXt101 & DeepLabv3+ & 0.895 & 0.951 \\
\midrule
\multirow{6}{*}{Al-Mg-Si SC}
 & ResNet50 & UNet & 0.285 & 0.294 \\
 & ResNet50 & DeepLabv3 & 0.398 & 0.476 \\
 & ResNet50 & DeepLabv3+ & 0.412 & 0.489 \\
 & SE-ResNeXt101 & UNet & 0.311 & 0.311 \\
 & SE-ResNeXt101 & DeepLabv3 & 0.463 & 0.545 \\
 & SE-ResNeXt101 & DeepLabv3+ & \textbf{0.472} & \textbf{0.550} \\
\midrule
\multirow{6}{*}{Al Pitting}
 & ResNet50 & UNet & 0.715 & 0.541 \\
 & ResNet50 & DeepLabv3 & 0.231 & 0.348 \\
 & ResNet50 & DeepLabv3+ & 0.253 & 0.374 \\
 & SE-ResNeXt101 & UNet & \textbf{0.752} & \textbf{0.570} \\
 & SE-ResNeXt101 & DeepLabv3 & 0.245 & 0.365 \\
 & SE-ResNeXt101 & DeepLabv3+ & 0.268 & 0.389 \\
\midrule
\multirow{6}{*}{Contact Corrosion}
 & ResNet50 & UNet & 0.923 & 0.911 \\
 & ResNet50 & DeepLabv3 & 0.842 & 0.834 \\
 & ResNet50 & DeepLabv3+ & 0.776 & 0.728 \\
 & SE-ResNeXt101 & UNet & \textbf{0.951} & \textbf{0.962} \\
 & SE-ResNeXt101 & DeepLabv3 & 0.856 & 0.914 \\
 & SE-ResNeXt101 & DeepLabv3+ & 0.526 & 0.628 \\
\midrule
\multirow{6}{*}{Carbon}
 & ResNet50 & UNet & 0.736 & 0.609 \\
 & ResNet50 & DeepLabv3 & 0.524 & 0.649 \\
 & ResNet50 & DeepLabv3+ & 0.586 & 0.672 \\
 & SE-ResNeXt101 & UNet & \textbf{0.762} & \textbf{0.621} \\
 & SE-ResNeXt101 & DeepLabv3 & 0.547 & 0.671 \\
 & SE-ResNeXt101 & DeepLabv3+ & 0.599 & 0.686 \\
\bottomrule
\end{tabular}
\end{table*}

\subsection{Out-of-Distribution for Quality Control}

\textbf{Why OOD?} Out-of-Distribution detection is a non-negotiable problem towards safe and deployable machine learning. It provides the first line of defense, preventing silent failures in deployed machine learning systems. It bounds AI capabilities by recognizing model knowledge limitations via training data, and allows safe fallback, enabling human oversight when needed. The term OOD detection first emerged in late-2016 with the seminal baseline work by \citet{hendrycks2017baseline}, which established that well-trained neural networks tend to assign higher softmax scores to in-distribution examples.

Density-based methods for OOD detection model the distribution of in-distribution data and flag low-density regions as OOD. These include approaches using variational autoencoders \citep{denouden2018improving}, normalizing flows \citep{kirichenko2020normalizing}, and more recently, diffusion models \citep{graham2023denoising}. \citet{morningstar2021density} provide analysis of when density estimation succeeds and fails for OOD detection. Compared to prior work in OOD literature \citep{yang2022openood, morningstar2021density, ganguly2024visual} Forte \citep{gangulyForteFindingOutliers2024a} presents distinct advantages.

\textbf{Why Forte?} Forte was chosen because it utilizes self-supervised representations to capture semantic features, built with a plug-and-play design - it works with any feature extractor, including intermediate layer features from material science specific neural networks. The per-point summary statistics incorporate elements of manifold estimation to account for local topology. This minimizes deployment overhead, eliminates additional model training requirements and provides zero-shot performance. Moreover, Forte requires no class labels, no exposure to OOD data during training and no restrictions to the architecture of predictive or generative models, with strong domain generalizations shown to synthetically generated images \citep{ganguly2025labeling}, MRI images \citep{gangulyForteFindingOutliers2024a}, HPC log anomaly detection \citep{chen2025k} and Large Language Model Guard-railing including jailbreak and toxicity detection\citep{ganguly2026trust}. This is enabled because the problem of out-of-distribution detection is very similar to algorithmic challenges such as anomaly detection, open-set recognition, novelty detection. Forte's underlying principle of typicality estimation also finds uses in syntactic typicality of neurosymbolic program synthesis \citep{ganguly2024proof}, as a proxy of uncertainty quantification  (UQ)\citep{ganguly2025grammars}.

\noindent
\textbf{AFM crystallites case study.}
We emphasize that quality control is not evaluated as a systematic or cross-domain study in this work.
Instead, we provide a focused case study on the AFM crystallites dataset to illustrate how out-of-distribution (OOD) detection can be used to surface low-quality segmentation results within a segmentation benchmarking pipeline.

In this setting, OOD samples correspond to segmentation outputs that deviate from the typical boundary morphology of crystalline spherulites.
Such deviations most commonly arise from incomplete, fragmented, or distorted crystallite boundaries, which are difficult to reliably detect using aggregate segmentation metrics alone.
While these predictions may still achieve moderately high IoU scores, they represent semantically poor segmentations from a materials characterization perspective.

\begin{figure*}[t]
    \centering
    \includegraphics[width=\textwidth]{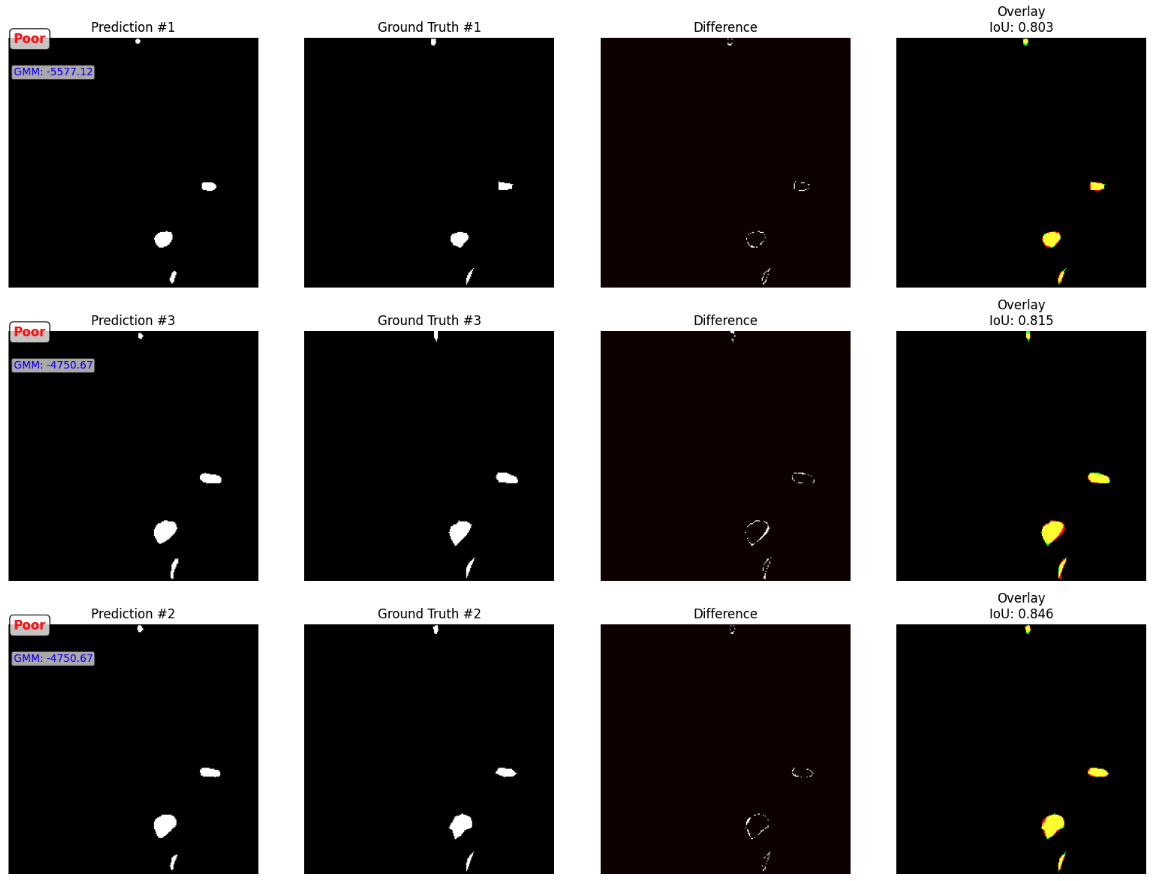}
    \caption{
    AFM crystallites segmentation failure cases flagged by OOD-based quality control.
    Each row shows a representative prediction assigned a low Forte score.
    From left to right: predicted segmentation, ground-truth annotation, pixel-wise difference, and overlay visualization with the corresponding IoU score.
    Although the IoU values remain moderately high, the predictions exhibit systematic errors in crystallite boundary delineation.
    These deviations from typical boundary morphology lead to low Forte scores, demonstrating the utility of OOD-based quality control for identifying subtle but semantically meaningful segmentation failures.
    }
    \label{fig:afm_ood_failure}
\end{figure*}

Forte assigns low typicality scores to these atypical outputs by comparing their latent representations against the distribution learned from in-distribution AFM samples.
As a result, segmentation failures associated with boundary inaccuracies are effectively flagged despite not being extreme outliers under standard performance metrics.
Representative OOD failure cases identified by this process are shown in Figure~\ref{fig:afm_ood_failure}.

\eat{
\textcolor{red}{@Mingjian: add atleast one plot/diagram showing how the segmentation quality degrades and how that reflects in a low Forte score. This is essential because quality control is a headline contribution. I would also link to this using a ref command in the main text of the paper such that the reviewers definitely read this. }

\textcolor{red}{@Mingjian: Some more information about the quality control/OOD task definition in this case and evaluation protocol would also be good. In your experience with what gets flagged via the pipeline. what is OOD in this case? are these images where there is a shift across modality? Shift across materials? Across labs/instruments? across time/batches? can you add some images over here to illustrate}
}

\subsection{Explainability and Interpretability}

The widespread adoption of deep learning in high-stakes domains has created an urgent need for explainable AI (XAI) methods that can illuminate the decision-making processes of complex models \citep{samek2021explaining, adadi2018peeking}. Explainability serves multiple purposes: building user trust, meeting regulatory requirements, debugging models, and enabling scientific discovery \citep{arrieta2020explainable}. XAI methods can be categorized along several dimensions \citep{nauta2023anecdotal, vilone2021notions}. The task of evaluating which explainability method is best suited given a problem statement is an active area of research \citep{ganguly2022machine}. 

While quality control identifies when predictions are unreliable (e.g., the boundary failures in Figure~\ref{fig:afm_ood_failure}), it does not reveal why valid decisions are made. To address this, we employ counterfactual analysis (Figure~\ref{fig:afm_case}d) to verify that the model attends to physically meaningful features, such as grain boundaries, rather than spurious correlations. Our choice of explainability method is guided by domain expert feedback on the qualitative visualizations.

\end{document}